\documentclass[runningheads]{llncs}
\usepackage[T1]{fontenc}
\usepackage{graphicx}
\usepackage{booktabs}
\usepackage[misc]{ifsym}

\usepackage{mwe}

\usepackage{float}
\usepackage{orcidlink}
\usepackage{subfig}
\usepackage{amsmath}
\usepackage{xcolor,colortbl}
\usepackage{xspace}
\usepackage{tikz}
\usepackage{colortbl}
\usepackage{wrapfig} 

\begin{document}

\title{Spatial Transfer Learning for Estimating PM$_{2.5}$ in Data-poor Regions }


\titlerunning{Spatial Transfer Learning for Estimating PM$_{2.5}$}


\author{Shrey Gupta\inst{1}\thanks{Both authors contributed equally to this work.} \orcidlink{0000-0002-0284-3162} (\Letter) \and
Yongbee Park\inst{4}$^*$ \orcidlink{0009-0004-7461-7668}  \and
Jianzhao Bi\inst{2} \orcidlink{0000-0003-3807-6927}\and
Suyash Gupta\inst{3} \orcidlink{0000-0002-3240-1840} \and
Andreas Züfle\inst{1} \orcidlink{0000-0001-7001-4123}\and
Avani Wildani\inst{1,5} \orcidlink{0000-0001-9457-8863} \and
Yang Liu\inst{1} \orcidlink{0000-0001-5477-2186} (\Letter)} 


\authorrunning{S. Gupta and Y. Park et al.}



\institute{Emory University, USA \\ 
\email{\{shrey.gupta,azufle,yang.liu\}@emory.edu} \and 
University of Washington, USA \\
\email{jbi6@uw.edu} \and 
University of California, Berkeley, USA \\
\email{suyash.gupta@berkeley.edu} \and 
Ingkle, South Korea \\
\email{yongbee.park@ingkle.com} \and
Cloudflare, USA \\
\email{agadani@gmail.com}}

\tocauthor{Shrey Gupta, Yongbee Park, Jianzhao Bi, Suyash Gupta, Andreas Züfle, Avani Wildani, Yang Liu}

\toctitle{Spatial Transfer Learning for Estimating PM$_{2.5}$ in Data-poor Regions}

\maketitle              

\begin{abstract}
Air pollution, especially particulate matter 2.5 (PM$_{2.5}$), is a pressing concern for public health
and is difficult to estimate in developing countries (\emph{data-poor} regions) due to a lack of ground sensors.  
Transfer learning models can be leveraged to solve this problem, as they use alternate data sources to gain knowledge (i.e., data from \emph{data-rich} regions).
However, current transfer learning methodologies do not account for dependencies between the source and the target domains.
We recognize this transfer problem as \emph{spatial} transfer learning and propose a new feature named \emph{Latent Dependency Factor} (LDF) that captures spatial and semantic dependencies of both domains and is subsequently added to the feature spaces of the domains. 
We generate LDF using a novel two-stage autoencoder model that learns from clusters of similar source and target domain data.
Our experiments show that transfer learning models using LDF have a $19.34\%$ improvement over the baselines. We additionally support our experiments with qualitative findings.

\newcommand{\customfootnote}[1]{%
  \begingroup
  \renewcommand{\thefootnote}{\fnsymbol{footnote}}%
  \footnotetext[2]{#1}%
  \endgroup
}
\customfootnote{Accepted for publication at ECML-PKDD 2024.}

\end{abstract}

\section{Introduction}
Air pollution, especially atmospheric aerosols smaller than 2.5 micrometers \textit{i.e.} PM$_{2.5}$ poses a significant concern to public health~\cite{sharma2020health}.
Emissions from vehicles~\cite{kinney2000airborne},
wildfires~\cite{candianwildfire}, and industrial processes~\cite{el2011primary} 
are major contributors to high PM$_{2.5}$ levels. 
Current approaches for measuring PM$_{2.5}$ involves using either remote sensing methodologies~\cite{bi2019impacts} 
or ground sensors~\cite{ayers1999teom}.
While satellite-based remote sensing methodologies are a low-cost way to measure PM$_{2.5}$, however, their data collection is affected by factors like cloudy weather and high surface reflectance, 
thereby significantly reducing the accuracy of measured PM$_{2.5}$ levels~\cite{bi2019impacts}.
Alternatively, installing PM$_{2.5}$ ground sensors yields highly accurate data as these sensors 
employ gravimetric data collection methodologies~\cite{ayers1999teom}. 
However, due to their high installation and maintenance costs~\cite{kumar2016join}, it is challenging to scale them in developing countries~\cite{dey2012variability}, creating an imbalance of \emph{data-rich} (developed) and \emph{data-poor} (developing) regions with PM$_{2.5}$ data for air pollution estimation.
  
Transfer learning (TL) can ameliorate this situation by utilizing \emph{data-rich} (source data) regions to learn a prediction model on \emph{data-poor} (target data) regions~\cite{pan2009survey}.
Prior research on estimating PM$_{2.5}$ through TL 
is geared towards time-series forecasting where the model learns historical data of an observed location (sensors) and forecasts the horizon (\textit{i.e.} future values) for the observed locations~\cite{fong2020predicting,ma2020air,yadav2022few,yao2024multi}.
Therefore, these models cannot estimate the PM$_{2.5}$ levels for locations where historical data is unavailable~\cite{vasiliev2020visualization}.
Alternatively, one can employ {\em Instance transfer learning} (ITL) models that avoid the limitations of time-series forecasting models by not relying on continuous temporal data~\cite{garcke2014importance,gupta2023boosting}.
ITL models reweigh source domain samples based on the target domain and subsequently combine the two domains. 

Unfortunately, ITL models are limited in estimating PM$_{2.5}$ as they overlook the {\em spatial and semantic correlations} 
in the datasets. 
PM$_{2.5}$ estimation data is uniquely heterogeneous and complex, containing topographical, meteorological, and geographical features.  These features exhibit \emph{spatial} autocorrelations (dependencies), \textit{i.e.} nearby locations tend to have similar PM$_{2.5}$ levels, as well as \emph{semantic} correlations (dependencies), \textit{e.g.} locations with similar meteorological and topographical conditions exhibit similar PM$_{2.5}$ levels with high likelihood~\cite{li2004similarity}. Spatial dependencies are prevalent within a domain, whereas semantic dependencies will likely arise when combining two domains (case for ITL).
We call this complex transfer problem as \emph{spatial} transfer learning.

In this paper, we solve \emph{spatial} transfer learning to improve PM$_{2.5}$ estimation by allowing source and target data points to learn from each other in the combined domain space.
We achieve this by introducing a new feature called {\em Latent Dependency Factor} (LDF) in both the source and target datasets to bridge the gap between the two domains. 
To generate LDF, we first learn a cluster of similar (spatially and semantically similar) data points for each sample, which are fed to our novel two-stage autoencoder model.
The first stage, \emph{encoder-decoder}, aims to learn a latent representation from the combined feature space of the cluster, 
while the second stage, \emph{encoder-estimator},  learns from the target label (PM$_{2.5}$ value). 
The LDF is highly correlated to the target (dependent) variable and contains learned dependencies from both domains.
To illustrate the benefits of LDF, we utilize real-world PM$_{2.5}$ data for the United States and Lima city in Peru. Our experiments include a comparative analysis of ML and TL models within the US boundaries, where we observe a $19.34\%$ improvement in prediction accuracy over baseline models. 
We also present a qualitative analysis showcasing the deployment of our methodology in \emph{data-poor} regions such as Lima. In summary, we make the following contributions:
\begin{enumerate}
    \item We present \emph{Latent Dependency Factor} (LDF), a new feature to learn the spatial and semantic dependencies within the combined source and target domains and close the gap between the two domains.
    
    \item We introduce a novel two-stage autoencoder model to generate LDF. It learns dependencies from the combined feature space of the clustered input data and the dependent variable.
    
    \item We explore the settings for \emph{spatial} transfer learning for PM$_{2.5}$ estimation in data-poor regions, a challenging problem with untrained test locations and sparse target and source locations causing minimal spatial autocorrelation. 
    
    \item We deploy our technique in Lima, Peru, and validate the results by domain experts due to the scarcity of true labels. This offers insights into the real-world application of our technique and its effectiveness.
    
\end{enumerate}

\section{Related Work}

\subsubsection{Estimating PM$_{2.5}$ via Transfer Learning}
Prior studies have utilized transfer learning for PM$_{2.5}$ estimation through time-series forecasting models, which learn from historical data of target sensors to forecast their future values. 
Fong et al.~\cite{fong2020predicting} incorporate Recurrent Neural Networks (RNN) in their temporal transfer learning model. 
Yao et al.~\cite{yao2024multi} employ Variational Auto-Encoders (VAE) using nearby sensors as source data, while Ma et al.~\cite{ma2020air} combine Long-Short Term Memory (LSTM) and RNN to forecast long-range PM$_{2.5}$ levels from short-range historical data. 
Yadav et al.~\cite{yadav2022few} leverage low-cost sensors as source data for temporal transfer learning for ground sensors. 
However, our problem is not suited for time-series forecasting due to missing temporal points and the lack of temporal matching between regions with varying meteorological conditions.

\subsubsection{Transfer Learning via Feature Augmentation}
Previous studies have improved model predictions by imputing features from another dataset~\cite{kumar2016join,liu2022feature} or generating synthetic samples to augment data~\cite{jaipuria2020deflating,vobecky2021artificial,veyseh2021augmenting}.
The former leverages datasets with low marginal distribution, while the latter focuses on augmenting samples rather than features. 
In the domain of transfer learning, Daume et al.~\cite{daume2007frustratingly} and Duan et al.~\cite{duan2012learning} introduce domain adaptation models — Feature Augmentation Method (FAM) and Heterogeneous Feature Augmentation (HFM), respectively — to create a common feature space using source and target features. 
These models are useful when the source and target domains have a dissimilar feature space, as noted by Pan et al.~\cite{pan2008transferring}, whereas our approach incorporates spatial and semantic dependencies during ITL for domains with similar feature spaces, high marginal distribution, and low spatial autocorrelation.

\section{Problem Formulation}
Our problem comprises the source region with higher PM$_{2.5}$ sensors and the target region with fewer sensors. The data is heterogeneous due to diverse features and complex due to spatial and semantic dependencies between its samples. 

Let $X_{f}^S$ be the feature set for the source domain with $m$ samples, and let $X_{f}^T$ be the feature set for the target domain with $n$ samples, such that $m >> n$, and contains $f$ features.
Let $Y^S$ and $Y^T$ be the source and target domain labels (PM$_{2.5}$ levels). 
Hence, $D^S = {(x_i^S, y_i^S)}_{i=1}^{m}$ is the source domain dataset, where $x_i^S \in X_{f}^S$ is the feature vector for the $i$-th PM$_{2.5}$ monitor, and $y_i^S \in Y_S$ is the corresponding PM$_{2.5}$ value at the sensor. 
Similarly, $D^T = {(x_i^T, y_i^T)}_{i=1}^{n}$ is the target domain dataset with $x_i^T$ and $y_i^T$ representing $i$-th monitor and its PM$_{2.5}$ value, respectively. 

\emph{Instance Transfer Learning} (ITL) methodologies are employed when the two domains have varying marginal distributions. They find a reweighing function $w(x)$ that adjusts the importance of each sample in the source domain based on its relevance to the target domain. The importance weights $w(x_{i}^S)$ are calculated for each sample $x_{i}^S$ in the source domain $D^S$, where $w(x_{i}^S)$ represents the degree of relevance of $x_{i}^S$ to the target domain $D^T$. This degree of relevance is often calculated using probability densities, expressed as $w(x_{i}^S) = \frac{P_{D^T}(x_i^S)}{P_{D^S}(x_i^S)}$, where $P_{D^T}(x_i^S)$  and $P_{D^S}(x_i^S)$ is the probability density of $x_i^S$ in the target domain and source domain respectively. The importance weights are applied to the source domain samples to obtain $\bar{D}_S = {(\bar{x}_i^S, y_i^S)}_{i=1}^{m}$ where $\bar{x}_i^S = w(x_{i}^S)\cdot x_{i}^S$. The reweighed source domain samples are used in the target domain for training; the combined domain is represented as $D^{\bar{S}T} = {(x_i^{\bar{S}T}, y_i^{\bar{S}T})}_{i=1}^{m+n}$.

\textbf{Our goal is to improve the estimation of PM$_{2.5}$, such that the combined domain $D^{\bar{S}T}$ after reweighing source domain data $D^S$ successfully captures the spatial and semantic dependencies.}

\section{Methodology}
\begin{figure*}[t]
  \centering
  \includegraphics[width=0.99\linewidth]{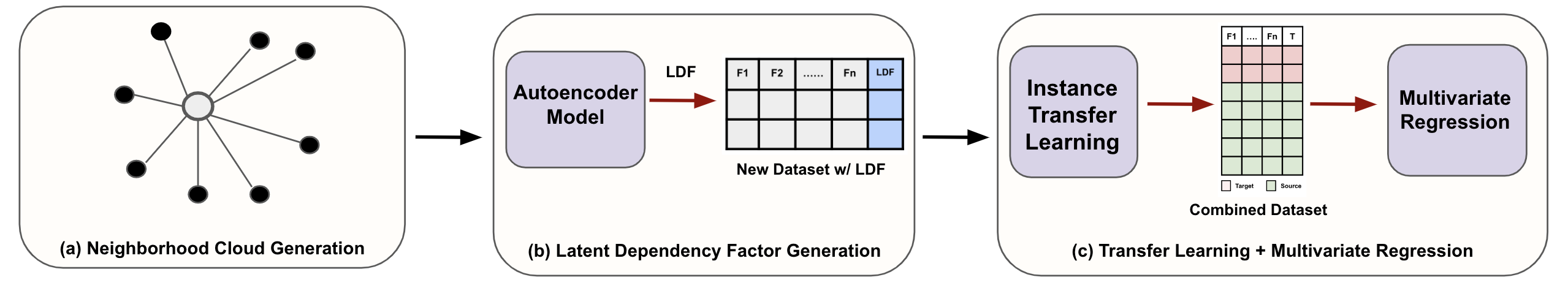}
  \caption{Framework for \emph{spatial} transfer learning via \emph{Latent Dependency Factor}}
    \label{methodology}
\end{figure*}

We introduce \emph{Latent Dependency Factor} (LDF), a new feature imputed in the dataset to achieve \emph{spatial} transfer learning for PM$_{2.5}$ estimation.
The LDF has the following attributes: (1) It is highly correlated to the observed variable (PM$_{2.5}$ value), (2) It captures the spatial dependencies (spatial autocorrelation between nearby locations), (3) It captures the semantic dependencies (semantic correlation in the combined data).

Imputing a new feature allows to learn a new loss function. Hence, if a function $f: X_{f}^{\bar{S}T} \rightarrow Y_T$ can predict the missing PM$_{2.5}$ values in the target domain $D_T$. 
Then, $f$ is learned by minimizing the empirical risk as, 

\begin{equation}
\label{eq1}
\min_{f} [\frac{1}{m+n}\sum_{i=1}^{m+n} \ell(y_i^{\bar{S}T}, f(x_i^{\bar{S}T})) + \lambda \cdot \Omega(f)]
\end{equation}

where $\ell(y, \hat{y})$ is the loss calculated between true PM$_{2.5}$ value ($y$) and predicted value ($\hat{y}$) (here $f(x_i^{\bar{S}T})$), $\Omega(f)$ is a regularization term, and $\lambda$ controls the trade-off between the empirical risk and model complexity. 
When a new feature is imputed, the empirical risk in (\ref{eq1}) is transformed as, 

\begin{equation}
\label{eq2}
\min_{f} [\frac{1}{m+n}\sum_{i=1}^{m+n} \widetilde{\ell}(y_i^{\bar{S}T}, \widetilde{f}(x_i^{\bar{S}T})) + \lambda \cdot \Omega(\widetilde{f})]
\end{equation}

with the new trained regressor, $\widetilde{f}$ and loss function $\widetilde{\ell}$. 
\emph{Hence, the new loss function allows obtaining a lower minimum.}
The framework for \emph{spatial} transfer learning via LDF contains 3 stages, as shown in Fig~\ref{methodology}, which we elaborate further.

\subsection{Neighborhood Cloud Generation}
The first stage (Fig~\ref{methodology}(a)) generates a neighborhood cloud of $k$ similar data points for each sensor in the source and target regions. This cloud is training data for the two-stage autoencoder model, allowing each sensor to learn the spatial dependencies of its neighbors and semantic dependencies between the two domains. The similarity between data points (sensors) is calculated by minimizing the $||L||_{2}$ distance across geographical, topographical, and meteorological features (see supplementary).

\subsection{Generating Latent Dependency Factor (LDF)}
\begin{figure*}[t]
  \centering
  \includegraphics[width=0.99\linewidth]{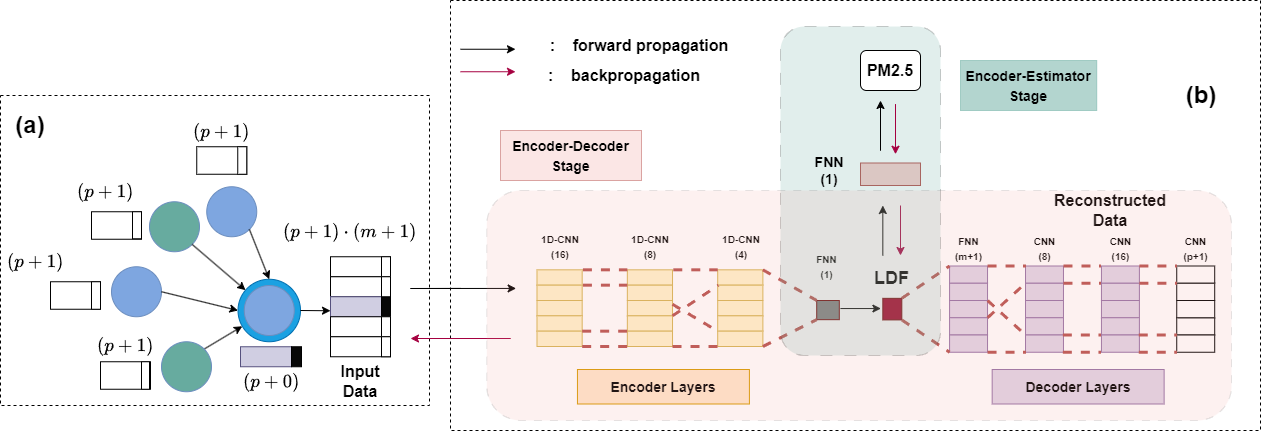}
  \caption{Two-stage autoencoder model for generating LDF.}
    \label{LCF-autoencoder}
\end{figure*}

After generating the neighborhood cloud, the subsequent steps involve generating the LDF, imputed as a new feature into the original dataset. This feature is derived using a two-stage autoencoder model (Fig.~\ref{LCF-autoencoder}(a)), where the input dataset (neighborhood cloud) utilizes features -- topographical, meteorological, geographical, and PM$_{2.5}$ levels.
We believe these predictors influence the PM$_{2.5}$ levels at the objective location (centroid of the cluster).
\emph{E.g.}, given a sensor location, $l_i$, in the target region, the predictors such as the \emph{wind-direction}, \emph{elevation}, \emph{population}, and more, for the surrounding sensors can influence the PM$_{2.5}$ levels at $l_i$ (spatial autocorrelation). 
Additionally, the sensor location, $l_i$, can be semantically correlated to another location, $l_j$, in the source region, influencing the PM$_{2.5}$ levels at $l_i$ in the combined dataset. 
In Fig.~\ref{LCF-autoencoder}(a), each sensor has $(p+1)$ features with $p$ features and a label. 
We first calculate the weight for each feature. This is achieved by finding the similarity (inverse distance) between the feature of the objective location and neighboring sensors.
This allows sensors with influential features to be given more importance.
Following the weighing, the features from $m$ sensors are stacked together with the objective location to generate the input data of size $(m+1) \cdot (p+1)$. 
The PM$_{2.5}$ for the objective location is voided by setting it to 0. This high-dimensional data is summarized into the LDF, using the two-stage autoencoder model shown in Fig.~\ref{LCF-autoencoder}(b).  

\paragraph{\textbf{Encoder-decoder Stage}}
The \emph{encoder-decoder} stage of the two-stage autoencoder model is similar to the standard autoencoder model, where the encoder first summarizes the input data to generate a latent value. The decoder employs backpropagation to train the autoencoder. The encoder and the decoder have three 1D-CNN layers with varying filter sizes, as shown in Fig.~\ref{LCF-autoencoder}(b). For the encoder, the kernel size of the first 2 CNN layers is chosen as 1 to achieve individual attention for each sensor and amplify the effectiveness of information summarization~\cite{kingma2018glow}. The third CNN layer has a kernel size 3 to retain the condensed pattern from multiple stations. Finally, the information is summed up using an FNN layer, which outputs the latent value, \emph{i.e.}, the LDF value.

\paragraph{\textbf{Encoder-estimator Stage}}
Since the input data consists of multiple features, we increase the attention on PM$_{2.5}$ labels using the \emph{encoder-estimator} stage. The estimator layer takes the encoded LDF value as input. It has a single FNN layer with a single weight and bias set. 
It utilizes back-propagation and PM$_{2.5}$ value of the objective location to train the encoder-decoder model and consequently optimize the LDF generation process. 
The autoencoder stages alternate training over the epochs.
We also explore extending LDF to include Aerosol Optical Depth (AOD)~\cite{sato1993stratospheric} feature, which we call LDF-A and which measures the aerial density of aerosols such as smoke, dust, and PM particles, in the \emph{encoder-esitmator} stage. 

\subsection{Transfer Learning and Multivariate Regression}
In Fig.\ref{methodology}(c), we employ Instance Transfer Learning (ITL) to mitigate discrepancies between source and target domain samples\cite{garcke2014importance}. This involves reweighing the source domain samples to align them closer to the target domain. The reweighed source data is combined with the target data, creating a unified dataset reflecting both domains' characteristics.

This combined dataset is subsequently used to train a multivariate regressor for predicting PM$_{2.5}$ values. The choice of regressor can range as \emph{polynomial-function} based, \emph{decision-tree} based, or \emph{ensemble} model. We employ an \emph{ensemble} regressor for our framework, given their high prediction accuracy~\cite{dong2020survey}.

\section{Evaluation}

\begin{figure*}[t]
  \centering
  \includegraphics[width=0.95\textwidth]{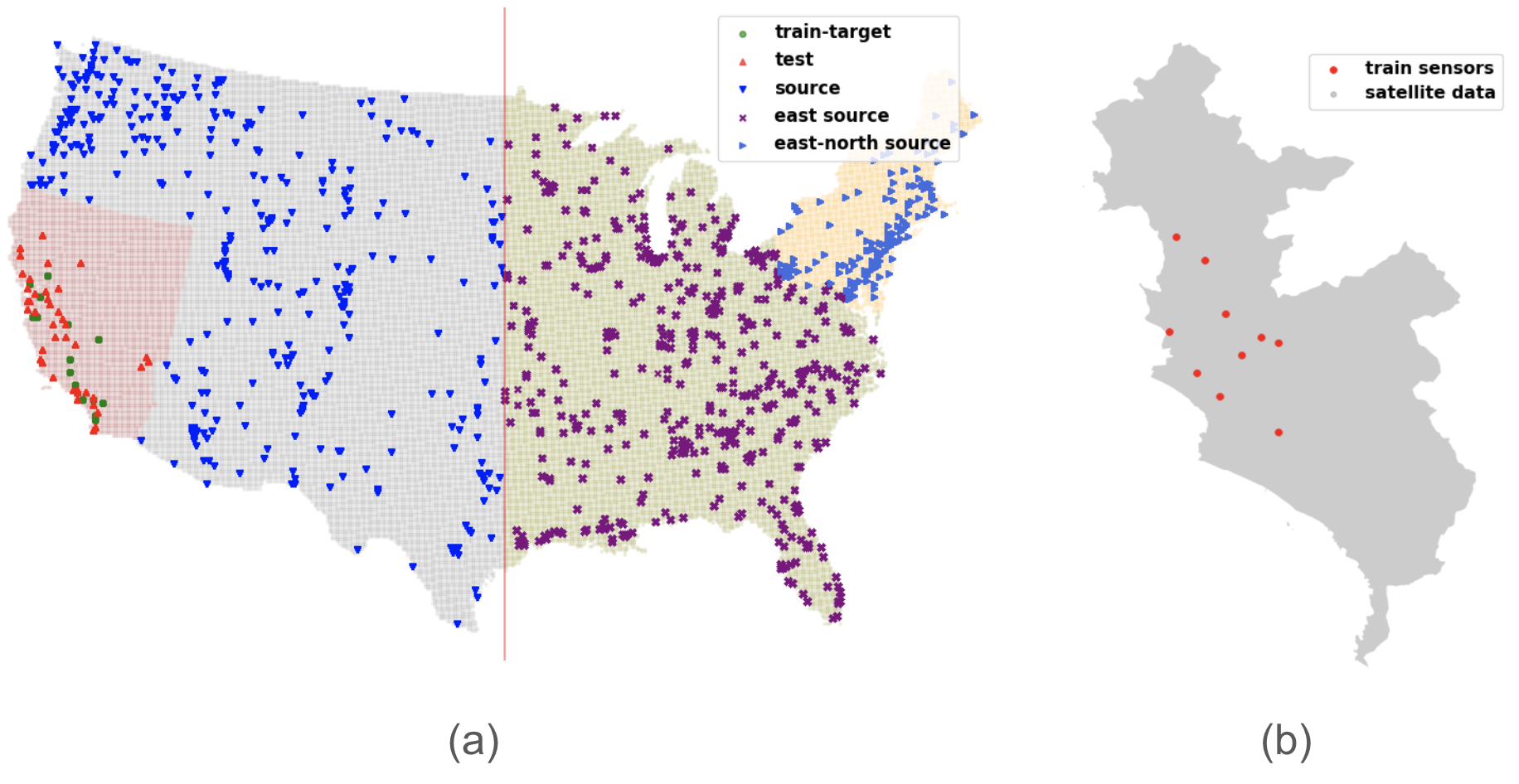}\vspace{-0.6cm}
  \caption{(a) US PM$_{2.5}$ ground sensors. The points in the pink target region represent sample training (green) and testing (red) sensors. The green and yellow regions represent the eastern and north-eastern source regions, respectively.
  (b) PM$_{2.5}$ sensors in Lima, Peru. Red points represent sensors used for training, and the grey area represents satellite data for testing.}
  \label{data-map}
\end{figure*}

\subsection{Datasets}
We employ existing PM$_{2.5}$ datasets of two distinct regions: the US~\cite{park2020estimating} and 
the Lima, Peru~\cite{vu2019developing}.
In comparison to other datasets~\cite{misc_beijing_pm2.5_381},
these corpora draw from diverse sources (EPA, NLDAS-2, and NED for the US and SENAMHI and JHU for Lima) 
and encompass a wide array of heterogeneous features such as \emph{wind patterns}, \emph{atmospheric pressure}, \emph{humidity levels}, \emph{potential energy} and more.

\subsubsection{United States dataset.}
As the US region has abundant PM$_{2.5}$ sensors, we select this dataset to simulate a transfer learning scenario within its geographical boundaries.
The US dataset has daily averaged PM$_{2.5}$ levels for 2011 using $1081$ sensors, as shown in Fig.~\ref{data-map}, with over 249k samples and 77 features. 
Although the sample size should be 1081$\times$365, some sensors were inactive on certain days (daily average active sensors: $\sim682$). 
This contributes to missing temporal points in the dataset, which limits the application of time-series forecasting methodologies.
We follow the prior work~\cite{park2020estimating} and use Layerwise Relevance Propagation~\cite{bach2015pixel} to extract 27 meteorological, topographical, and geographical features.
As illustrated in Fig~\ref{data-map}(a), 
we select two source regions, the eastern US (highlighted green; marker: \textbf{x}) and north-eastern US (highlighted yellow; marker: \begin{tikzpicture}[scale=0.2]
\fill[black] (0,0) -- (1,0.5) -- (0,1) -- cycle;
\end{tikzpicture}
) 
and a target region, California-Nevada (highlighted pink).
Prior works~\cite{bi2020incorporating} show that the California-Nevada region has a diverse landscape compared to the remaining US, thereby simulating a TL scenario with distribution shift and low spatial correlation among the two domains.

We sample the 128 target region sensors into sets of 5, 7, 9, and 11 sensors 
to have fewer samples. The remaining sensors are used for testing.
For cross-validation (CV), we use 20 random samples per sensor.
We extrapolate the active sensors per day and generate a neighborhood cloud for each sensor that includes both source and target sensors.
Next, the clustered data is used to generate the LDF which is fed to the transfer models.
Our reported R$^{2}$ and RMSE values represent averages across the 20 CVs. 
The features are normalized before model training.
For qualitative analysis (Section~\ref{ss:qualitative-analysis}), 
we use $\sim19.5$ million unlabeled satellite data samples from the California-Nevada region.

\subsubsection{Lima dataset}
Given the dearth of sensors in the Lima data, it is a use-case of real-world transfer learning, where the source data is the complete US dataset (249k+ samples, 27 features).
Lima region has 10 PM$_{2.5}$ sensors, as shown in Fig~\ref{data-map}(b), with 2419 samples and 21 features for the year 2016. 
Lima and the US datasets have only 14 common features
(see supplementary material).
For the qualitative analysis, the Lima satellite data contains 5959 samples covering the entire Lima region, as shown in Fig~\ref{data-map}(b) (highlighted grey).
We use all 10 sensors and the US dataset to construct the neighborhood cloud data such that each \emph{day of the year (doy)} between the two datasets (e.g., day 17 in Lima matched with day 17 in the US) are extrapolated to generate the clusters.
However, the matching of \emph{doy} is not on the same year or season between the two domains to have a real-world transfer condition with minimal alignment.

\subsection{Prediction Models}
\subsubsection{Machine Learning (ML) Models.}
    We select two popular ML models, \textbf{Random Forest Regressor (RF)}~\cite{hu2017estimating} and \textbf{Gradient Boosting Regressor (GBR)} \cite{zhang2021satellite},
    trained on only the target region data and tested on the remaining test data.
    The RF and GBR have parameters varied as \emph{n-estimators:} \{100, 400, 1000\}, \emph{max-depth}: \{4, 8, inf\} with \emph{max-leaf-node}: \{4, 8, inf\} for RF and \emph{learning-rate}: \{0.1, 0.5, 1.0\} for GBR, to get the best fit.

\subsubsection{Transfer Learning (TL) Models}
We select competitive ITL models~\cite{loog2012nearest,sugiyama2007direct,huang2006correcting} for the regression task and train them on target and source region data.
\begin{enumerate}
        \item \textbf{Nearest Neighbor Weighing (NNW)}: The NNW~\cite{loog2012nearest} reweighs the source samples by creating a Voronoi tesselation for each sample and counts the number of target samples falling inside it. 
        The model parameters are varied as: \emph{neighbors}: \{6, 8, 10\} and \emph{n-estimator}: Decision Tree Regressor with \emph{depth}: \{6, 8, inf\} to get the best fit.
        \item \textbf{Kullback–Leibler Importance Estimation Procedure (KLIEP)}: The KLIEP reweighs the source samples to minimize the KL divergence between the source and target domains~\cite{sugiyama2007direct}.
        The model parameters are varied as: \emph{kernel}: \{rbf, poly\}, \emph{gamma}: \{0.1, 0.5, 1.0\}, and \emph{n-estimator}: Decision Tree Regressor with \emph{depth}: \{6, 8, inf\} to get the best fit.
        \item \textbf{Kernel Mean Matching (KMM)}: The KMM reweighs the source samples such that means of source and target samples in reproducible kernel Hilbert space is minimized~\cite{huang2006correcting}.
        The model parameters are varied as: \emph{kernel}: \{rbf, poly\}, \emph{gamma}: \{0.1, 0.5, 1.0\}, and \emph{n-estimator}: Decision Tree Regressor with \emph{depth}: \{6, 8, inf\} to get the best fit.
        \item \textbf{Fully-connected Neural Network (FNN)}: 
        The FNN transfer model, although not an ITL model, is utilized to validate the performance of non-ITL models on the PM$_{2.5}$ data.
        It uses 3 fully connected layers: \emph{nodes}: 128, \emph{activation-function}: Relu, and 1 final layer with a single node and a linear activation function. 
        It was trained on LDF-imputed source data and transferred by fine-tuning over LDF-imputed target data.
\end{enumerate}

The TL models are trained on data sans LDF, LDF, and LDF-A-imputed data.
We use the GBR model as the multivariate regressor to predict PM$_{2.5}$, with parameters varied as: \emph{estimators:} \{100, 400, 1000\}, \emph{max-depth}: \{4, 8, inf\}, \emph{max-leaf-node}: \{4, 8, inf\}, and \emph{learning-rate}: \{0.1, 0.5, 1.0\} to get the best fit.
The source code, datasets, and final hyperparameter values are available at:

\noindent\textit{\url{https://github.com/YongbeeIngkle/spatial-transfer-learning.git}}.

\subsection{Optimal $k$ for Neighborhood Cloud}
In Fig.~\ref{combined-ablation}(a), we use the eastern US as source data and vary the size of the neighborhood cloud ($k$) for the NNW [LDF] model as \{4, 8, 12, 16\}.
Our choice of $k$ mimicked optimizing parameters, ceasing at 16 due to high computational costs.
We observe that $k=4$ has the worst performance, while for the remaining values, there is no observable difference for sensors $\geq 9$. 
For sensors $\leq 9$, $k=12$ has the most optimal performance.
Hence, we chose $k=12$ to optimize the computation and generalizability of the model.

\begin{table*}
    \centering
    \caption{Source: Eastern US (best highlighted; second-best underlined)}
    \label{east-source}
    \small
    \begin{tabular}{|>{\columncolor{gray!30}}c|*{1}{c}*{1}{c|}*{1}{c}*{1}{c|}*{1}{c}*{1}{c|}*{1}{c}*{1}{c|}}
        \hline
        & \multicolumn{8}{c|}{\textbf{Sensors}}\\
        \cline{2-9}
        & \multicolumn{2}{c|}{\textbf{5}}
                    & \multicolumn{2}{c|}{\textbf{7}}
                        & \multicolumn{2}{c|}{\textbf{9}}
                            & \multicolumn{2}{c|}{\textbf{11}}\\
        \cline{2-9}
        \textbf{Model} & \textbf{R$^{2}$} & \textbf{RMSE} 
         & \textbf{R$^{2}$} & \textbf{RMSE} 
         & \textbf{R$^{2}$} & \textbf{RMSE} 
         & \textbf{R$^{2}$} & \textbf{RMSE} \\
        \hline
        \hline
        RF & -0.082& 8.855& 0.002& 8.565& 0.066& 8.387& 0.071& 8.311\\
        
        GBR & -0.061& 8.684& 0.064& 8.210& 0.177& 7.857& 0.157& 7.891\\
        
        NNW & 0.236& 7.563& 0.263& 7.447& 0.280& 7.406& 0.296& 7.288\\
        
        KLIEP & 0.155& 7.960& 0.192& 7.801& 0.200& 7.811& 0.222& 7.666\\

        KMM & 0.197& 7.757& 0.226& 7.634& 0.242& 7.601& 0.258& 7.479\\

        FNN & -0.064& 8.818& -0.350& 9.715& 0.009& 8.629& -0.039& 8.765\\
        
        \cline{1-9} 
        
        NNW [LDF] & \textbf{0.247}& \textbf{7.494}& \textbf{0.336}& \textbf{7.061}& \textbf{0.378}& \textbf{6.874}& \textbf{0.378}& \textbf{6.838}\\
        
        NNW [LDF-A] & 0.225& 7.596& 0.298& 7.230& \underline{0.359}& \underline{6.973}& \underline{0.359}& \underline{6.924}\\
                
        KLIEP [LDF] & 0.202& 7.724& 0.278& 7.370& 0.325& 7.173& 0.336& 7.073\\
        
        KLIEP [LDF-A] & \underline{0.232}& \underline{7.584}& 0.267& 7.427& 0.319& 7.201& 0.330& 7.100\\
        
        KMM [LDF] & 0.210& 7.671& \underline{0.302}& \underline{7.236}& 0.353& 7.013& 0.352& 6.971\\
        
        KMM [LDF-A] & 0.196& 7.723& 0.295& 7.277& 0.330& 7.134& 0.333& 7.067\\

        FNN [LDF] & -0.255& 9.532& -0.141& 9.082& 0.072& 8.374& 0.087& 8.236\\

        FNN [LDF-A] & -0.150& 9.146& -0.105& 8.990& 0.091& 8.275& 0.078& 8.287\\
        
        \hline
    \end{tabular}
\end{table*}

\subsection{Results and Analysis}
In Table~\ref{east-source} and Table~\ref{north-east-source}, we compare the performance of various models with the eastern US and the north-eastern US as source datasets, respectively.

\paragraph{\noindent\textbf{Eastern US as Source Data.}}
First, we compare the ML and TL sans LDF models. 
In Table~\ref{east-source}, we observe that NNW, KLIEP, and KMM have a positive transfer (improved accuracy), with NNW having the best performance. We observe an unpredictable performance for the FNN transfer model, validating that non-ITL models are less suited for such transfer problems.
Next, we illustrate the impact of the \emph{Latent Dependency Factor} (LDF) on TL models. 
We observe an improvement in estimation accuracy for NNW, KLIEP, and KMM (for both LDF and LDF-A), where NNW [LDF] is the best-performing model. 
For the FNN model, LDF has no notable effect as it caters to only ITL models.
The high performance of NNW is due to the Voronoi tesselation neighborhood it uses for reweighing source samples. This allows it to capture similar data points in its neighbor, a spatially preferred reweighing for the PM$_{2.5}$ data.

\paragraph{\noindent\textbf{North-eastern US as Source Data.}}
In Table~\ref{north-east-source}, we observe a positive transfer for NNW and KLIEP models, with NNW having the best performance. 
KMM shows a negative transfer~\cite{rosenstein2005transfer} due to the high marginal distribution present between the target and source datasets~\cite{ito2004spatial}; unable to be minimized in reproducing kernel Hilbert space (RKHS)~\cite{huang2006correcting}.
Like earlier, the FNN transfer model has an unpredictable performance. 
When the LDF is introduced, we observe an improvement in estimation accuracy for NNW and KLIEP models. NNW [LDF] and NNW [LDF-A] are the best-performing models. KMM [LDF-A] shows improvement for more sensors ($\geq$11).
As expected, the FNN models using LDF and LDF-A show no improvement.

\begin{table*}
    \centering
    \caption{Source: North Eastern US (best highlighted; second-best: underlined)}
    \label{north-east-source}
    \small
    \begin{tabular}{|>{\columncolor{gray!30}}c|*{1}{c}*{1}{c|}*{1}{c}*{1}{c|}*{1}{c}*{1}{c|}*{1}{c}*{1}{c|}}
        \hline
        & \multicolumn{8}{c|}{\textbf{Sensors}}\\
        \cline{2-9}
        & \multicolumn{2}{c|}{\textbf{5}}
                    & \multicolumn{2}{c|}{\textbf{7}}
                        & \multicolumn{2}{c|}{\textbf{9}}
                            & \multicolumn{2}{c|}{\textbf{11}}\\
        \cline{2-9}
         \textbf{Model} & \textbf{R$^{2}$} & \textbf{RMSE} 
         & \textbf{R$^{2}$} & \textbf{RMSE} 
         & \textbf{R$^{2}$} & \textbf{RMSE} 
         & \textbf{R$^{2}$} & \textbf{RMSE} \\
        \hline
        \hline
        RF & -0.082& 8.855& 0.002& 8.565& 0.066& 8.387& 0.071& 8.311\\
        
        GBR & -0.061& 8.684& 0.064& 8.210& 0.177& 7.857& 0.157& 7.891\\
        
        NNW & 0.199& 7.732& 0.294& 7.286& 0.301& 7.297& 0.298& 7.257\\
        
        KLIEP & 0.098& 8.180& 0.219& 7.650& 0.263& 7.494& 0.270& 7.408\\

        KMM & -0.142& 9.053& -0.070& 8.809& 0.232& 7.640& 0.246& 7.526\\

        FNN & 0.022& 8.448& -0.006& 8.598& 0.091& 8.266& 0.078& 8.307\\
        
        \cline{1-9}
        
        NNW [LDF] & \textbf{0.225}& \textbf{7.592}& \underline{0.317}& \underline{7.157}& \underline{0.376}& \underline{6.886}& \textbf{0.392}& \textbf{6.751}\\
        
        NNW [LDF-A] & \underline{0.201}& \underline{7.702}& \textbf{0.320}& \textbf{7.122}& \textbf{0.378}& \textbf{6.873}& \underline{0.374}& \underline{6.847}\\
                
        KLIEP [LDF] & 0.164& 7.889& 0.275& 7.363& 0.353& 7.011& 0.360& 6.924\\
        
        KLIEP [LDF-A] & 0.170& 7.860& 0.270& 7.396& 0.342& 7.068& 0.348& 6.991\\
        
        KMM [LDF] & -0.265& 9.409& 0.009& 8.468& 0.188& 7.749& 0.257& 7.389\\
        
        KMM [LDF-A] & -0.152& 9.042& -0.029& 8.566& 0.172& 7.845& 0.288& 7.260\\

        FNN [LDF] & 0.036& 8.429& -0.052& 8.761& 0.131& 8.061& 0.237& 7.566\\

        FNN [LDF-A] & -0.060& 8.774& 0.045& 8.390& 0.159& 7.983& 0.207& 7.708\\
        
        \hline
    \end{tabular}
\end{table*}

\subsection{Qualitative Analysis}
\label{ss:qualitative-analysis}

\begin{figure*}[t]
  \centering
  \includegraphics[width=0.99\linewidth]{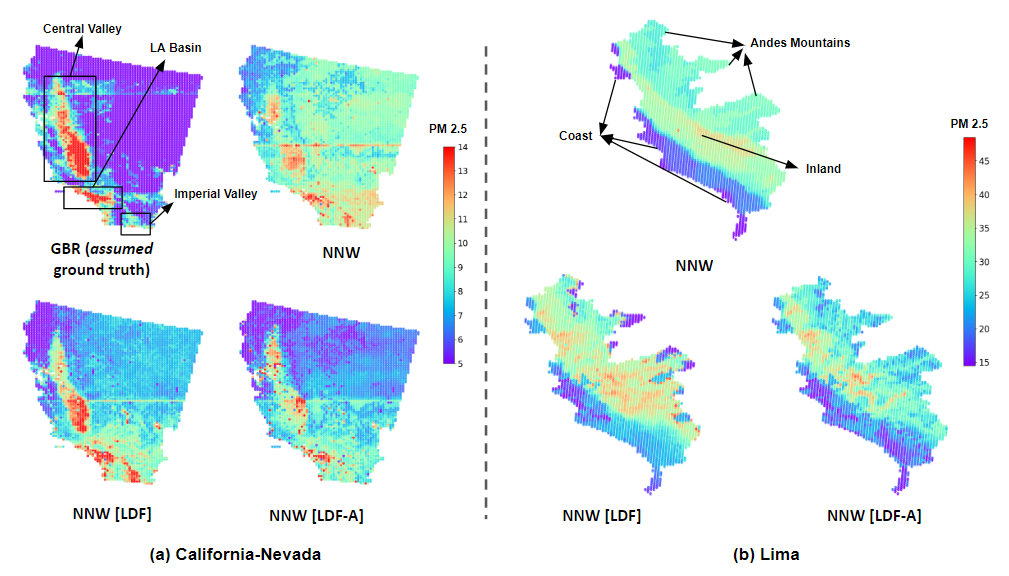}
  \caption{(a) Annual mean PM$_{2.5}$ prediction for \emph{California-Nevada},  trained using GBR and NNW with and without LDF features (9 sensors). (b) Annual mean PM$_{2.5}$ prediction for Lima region trained using NNW models.} 
    \label{map-prediction-plot}
\end{figure*}

While improving prediction accuracy is crucial, visualizing PM$_{2.5}$ patterns on geo-maps is also valuable. 
We visualize PM$_{2.5}$ estimations for the California-Nevada region and the Lima, Peru region in Fig.~\ref{map-prediction-plot}(a) and Fig.~\ref{map-prediction-plot}(b), respectively. 
For this analysis, we need a ground truth against which all the models can be compared. 
We use the GBR model, trained on all 128 monitors (249k+ samples) and estimated on the unlabeled satellite data ($\sim19.5$ M samples), 
and use its predicted geo-map as the assumed ground truth for verification.
We use 9 sensors and the eastern US as source data for transfer models (NNW, NNW[LDF], NNW[LDF-A]).
For Lima, we use all 10 sensors and the eastern US as the source data for the three transfer models. 
The true labels for Lima were unavailable, and domain experts (environmental scientists) were consulted for analysis.

Due to the scarcity of target domain data, this qualitative analysis aims to observe if transfer models successfully capture glaring PM$_{2.5}$ estimation patterns.

\subsubsection{California-Nevada Region.}
In Fig.~\ref{map-prediction-plot}(a), we observe that the NNW [LDF] model has the most accurate PM$_{2.5}$ estimation in the hotspots (solid boxes in the GBR map). It accurately captures patterns in the \emph{Central Valley} and the \emph{Los Angeles Basin} but overestimates in the \emph{Imperial Valley}. NNW [LDF-A] has the second-best performance but has a patchy estimation in the \emph{Central Valley}. For NNW, we observe obscure patterns that are patchy and underestimated in the \emph{Central Valley} and highly overestimated in the \emph{Imperial Valley}. 

\subsubsection{Lima Region.}
In Fig.~\ref{map-prediction-plot}(b), all models exhibit lower PM$_{2.5}$ levels near the \emph{coast} and higher levels moving \emph{inland}, a pattern validated by domain experts.
However, NNW [LDF] has a clearer concentration gradient of \emph{inland} PM$_{2.5}$ compared to the other models. 
Near the \emph{Andes mountain ranges}, the PM$_{2.5}$ is the lowest, which the NNW[LDF] model accurately captures but slightly and highly overestimated by the NNW [LDF-A] and NNW models, respectively. 
These observations confirm the improvement of prediction by LDF-based TL models.

Additionally, we performed a 60:40 train-test split on the Lima sensors and trained NNW and NNW [LDF] TL models using 3-fold cross-validation, with the complete US as the source data. The results for [R$^{2}$, RMSE] for NNW and NNW[LDF], respectively, were [0.476, 9.852] and [0.558, 9.091]. Hence, NNW [LDF] outperforms NNW, validating the quantitative analysis.

\subsection{Ablation Study}

\begin{figure*}[t]
  \centering
  \includegraphics[width=\linewidth]{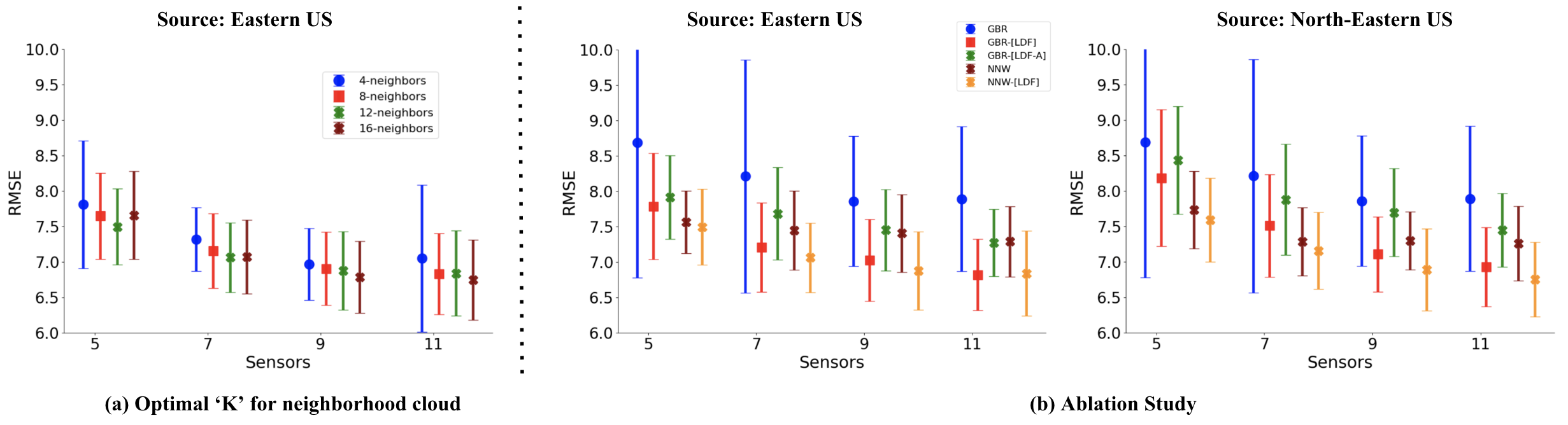}
  \caption{(a) Comparing performance of NNW [LDF] model when neighborhood cloud uses k = \{4, 8, 12, 16\} neighbors. (b) Ablation study comparing the performance of GBR, GBR [LDF], GBR [LDF-A], NNW, and NNW [LDF] models.}
    \label{combined-ablation}
\end{figure*}

For the ablation study, we use GBR instead of ITL models to validate the performance of non-transfer models using LDF-imputed data. 
Fig.\ref{combined-ablation}(b) shows the comparison between GBR [LDF], GBR [LDF-A], GBR (target only), NNW, and NNW [LDF]. 
For both the eastern US and the north-eastern US as source data, GBR [LDF] is the second-best performing model. 
Though it doesn't outperform NNW [LDF], the improved predictions highlight LDF's effectiveness.

The performance of FNN [LDF] and FNN [LDF-A] in Table~\ref{east-source} and Table~\ref{north-east-source} further tests LDF with non-ITL models, confirming that LDF is effective with ITL and multivariate regression models but not other transfer models.

\section{Discussion}
While the evaluation results show the improvement using the LDF, we further analyze the correlation between LDF and PM${2.5}$, as shown in Table~\ref{corr-table}, where LDF demonstrates the highest correlation with the dependent variable, indicating strong predictive power and feature importance~\cite{hall1999correlation}. This experiment uses an LDF-imputed dataset of 10 target sensors and eastern US source data.

\begin{table*}
    \centering
    \caption{Most correlated features (5) to PM$_{2.5}$ variable.}
    \label{corr-table}
    \small
    \begin{tabular}{|>{\columncolor{gray!30}}c|c c c c c|}
        \hline
        \rowcolor{gray!30} \textbf{Method} & \textbf{LDF} & \textbf{Pressfc} & \textbf{Dswrfsfc} & \textbf{Elev} & \textbf{Ugrd10m} \\ \hline
        \emph{Corr Coeff} & 0.754 & 0.208 & 0.181 & 0.179 & 0.156 \\      
        \hline
    \end{tabular}
\end{table*}

\textbf{Deployment in Lima:} 
Despite the lack of ground labels for deploying the LDF-based NNW model in Lima, Peru, it is important to address the pressing issue: Lima is the second most polluted city in the Americas~\cite{tapia2020pm} and suffers from a scarcity of sensors~\cite{vu2019developing} (Peru is a developing country). Our model provides a groundbreaking outcome in PM$_{2.5}$ estimation for Lima and serves as a vital first step toward implementing similar models in other \emph{data-poor} regions.

\subsection{Limitations and Future Work}
While our methodology improves PM$_{2.5}$ estimation, further exploration, and alternate improvements are still needed, which we outline below.

\subsubsection{Experiments with alternate datasets}
Previous experiments with the US and Lima data are comprehensive but do not include datasets lacking spatial and semantic dependencies~\cite{misc_beijing_pm2.5_381}. This was done primarily to ensure accurate and comprehensive data for modeling and estimation. Future plans include expanding our study to incorporate such datasets.

\subsubsection{Capturing temporal trends}
The LDF feature captures spatial and semantic dependencies but lacks focus on temporal trends in the data due to missing temporal points. In the future, we aim to extend this technique to time-series data, aiming for prediction rather than forecasting~\cite{zheng2008transferring}.

\subsubsection{Extending to alternate domains}
While our focus lies in PM$_{2.5}$ estimation, testing the LDF on alternate domains like wildfire estimation and weather forecasting is useful due to the presence of similar spatial patterns. Future studies should explore these applications and develop new LDF features accordingly.

    
    


\section{Conclusion}
This paper addresses the problem of \emph{spatial} transfer learning for estimating PM$_{2.5}$ levels, emphasizing transfer between regions with low autocorrelation and predicting at unseen test locations. 
We aim to improve \emph{instance transfer learning} (ITL) models, which often overlook spatial and semantic dependencies in the data. 
We introduce the \emph{Latent Dependency Factor} (LDF) to capture these dependencies, integrating it as a new feature in both source and target datasets. 
Our experiments on US and Peru datasets demonstrate LDF's effectiveness in improving PM${2.5}$ estimation. 
Furthermore, qualitative analysis of these datasets confirms that the LDF captures larger PM${2.5}$ patterns missed by regular transfer models. 
While more future work remains in this space, we believe our approach of achieving \emph{spatial} transfer learning using \emph{Latent Dependency Factor} is a promising and novel solution for this highly complex domain.

\end{document}